\documentclass[runningheads]{llncs}
\usepackage{graphicx}
\usepackage{hyperref}

\usepackage{comment}
\usepackage{multirow}
\usepackage{color}
\usepackage{tikz}
\def\checkmark{\tikz\fill[scale=0.4](0,.35) -- (.25,0) -- (1,.7) -- (.25,.15) -- cycle;} 

\begin{document}
\title{Deep learning for scene recognition from visual data: a survey}
\titlerunning{Deep learning for scene recognition: a survey}
%
\author{Alina Matei\inst{1,\thanks{Both authors contributed equally to this study}}\and
Andreea Glavan\inst{1,*} \and Estefan\'ia Talavera\inst{1}\orcidID{0000-0001-5918-8990}}%
\authorrunning{Talavera et al.}
%
\institute{Bernoulli Institute for Mathematics, Computer Science and Artificial Intelligence, University of Groningen, Nijenborgh 9, 9747 AG, Groningen, The Netherlands\\
\email{e.talavera.martinez@rug.nl}}
\maketitle              
\begin{abstract} 
The use of deep learning techniques has exploded during the last few years, resulting in a direct contribution to the field of artificial intelligence. This work aims to be a review of the state-of-the-art in scene recognition with deep learning models from visual data. Scene recognition is still an emerging field in computer vision, which has been addressed from a single image and dynamic image perspective. We first give an overview of available datasets for image and video scene recognition. Later, we describe ensemble techniques introduced by research papers in the field. Finally, we give some remarks on our findings and discuss what we consider challenges in the field and future lines of research. This paper aims to be a future guide for model selection for the task of scene recognition. 
\end{abstract}

\section{Introduction}

Recognizing scenes is a task that humans do on a daily basis. When walking down the street and going from one location to the other, tends to be easy for a human to identify where s/he is located. During the past years, deep learning architectures, such as Convolutional Neural Networks (CNNs) have outperformed traditional methods in many classification tasks. These models have shown to achieve high classification performance when large and variety datasets are available for training. Nowadays, the available visual data is not only presented in a static format, as an image, but also in a dynamic format, as video recordings. The analysis of videos adds an additional level of complexity since the inherent temporal aspect of video recordings must be considered: a video can capture scenes which suffer temporal alterations. Scene recognition with deep learning has been addressed by ensemble techniques that combine different levels of semantics extracted from the images, e.g. recognized objects, global information, and context at different scales.  

Developing robust and reliable models for the automatic recognition of scenes is of importance in the field of intelligent systems and artificial intelligence since it directly supports real-life applications. For instance, \textit{Scene and event recognition} has been previously addressed in the literature \cite{event_albums,objscene}. \textit{Scene recognition for robot localization} with indoor localization for mobile robots is one of the emerging application scopes of scene recognition \cite{robots_stateoftheart,robots_maps,robots_vgg_2}. According to the authors of \cite{robots_vgg_2}, in the following two decades, every household could own a social robot employed for housekeeping, surveillance or companionship tasks. In the field of lifelogging, collections of photo-sequences have proven to be a rich tool for the understanding of the behaviour of people. In \cite{furnari_video,hierarchical_food_scenes} methods were develop for the analysis of egocentric image collected by wearable cameras. The above-mentioned approaches address the recognition of scenes either following  an image-based approach or a video or photo-sequence based approach.

As contributions, (i) to the best of our knowledge, this is the first survey that collects works that address the task of scene recognition with deep deep learning from visual data, both from images and videos. Moreover, (ii) we describe available datasets which assisted the fast 
advancement in the field.

This paper is structured as follows: in Section \ref{sec:datasets} we discuss the available datasets supporting scene and object focused recognition. Section \ref{sec:methods} addresses the methodology of the state-of-the-art techniques and approaches discussed in the paper at hand. Furthermore, in Section \ref{sec:discussion} we discuss the presented approaches. Finally, in Section \ref{sec:conclusion} we draw some conclusions.

\section{Datasets for scene recognition} \label{sec:datasets}

The latest advancements in deep learning methods for scene recognition are motivated by the availability of large and exhaustive datasets and hardware that allows the training of deep networks. Thus, deep learning CNNs are applied to tackle the complexity and high variance of the task of scene recognition. 

The inherent difficulty of scene recognition is related to the nature of the images depicting a scene context. Two major challenges were described in \cite{scene_recognition_challenges}: 
\begin{itemize}
    \item \textit{Visual inconsistency} refers to low inter-class variance. Some scene categories can share similar visual appearances which create the issue of class overlaps. Since images belonging to two different classes can be easily confused with one another, the class overlap cannot be neglected. 

    \item \textit{Annotation ambiguity} describes a high intra-class variance of scene categories. Demarcation of the categories is a subjective process which is highly dependent on the experience of the annotators, therefore images from the same category can showcase significant differences in appearance.
\end{itemize}

The majority of the available datasets are focused on object categories providing labels \cite{imagenet,caltech256,cifar,coil100}, bounding boxes \cite{open_images} or segmentations \cite{open_images,coco}. ImageNet \cite{imagenet}, COCO (Common Objects in Context)\cite{coco}, and Open Images \cite{open_images} are well known in the field of object recognition. Even though these dataset were  built for object recognition, transfer learning has shown to be an effective approach when aiming to apply them for scene recognition. 

\begin{figure}[h!]
 \caption{Example of samples of the publicly available datasets as described in Table \ref{tab1}. Samples are presented from the same classes amongst similar datasets (i.e. scene, video and object centric) in order to emphasize the diversity of the image and video data. For the video-centric datasets (i.e. Maryland "in-the-wild", YUPENN, YUP++) representative video frames are presented.}
    \centering
    \includegraphics[width=\columnwidth]{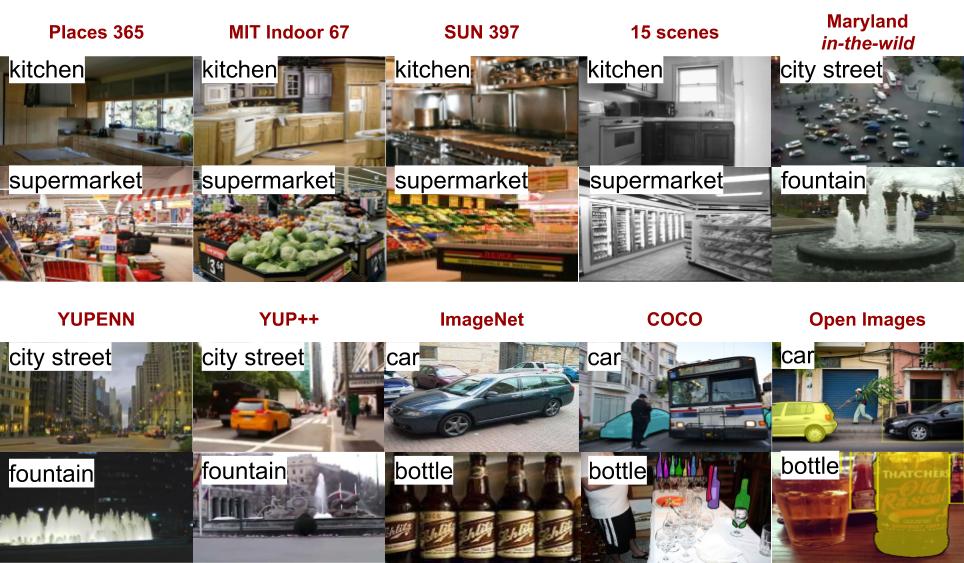}
    \label{fig:places}
\end{figure}

In the literature we can find the 15-scenes \cite{15_scenes}, MIT Indoor67 \cite{mitindoor67}, SUN397 \cite{sun397}, and Places365 \cite{places365} as  scene-centered datasets. More specifically, the Places project introduced Places365 as a reference dataset, which is composed of 434 scenes which account for 98\% of the type of scenes a person can encounter in the natural and man-made world.
A total of 10 million images were gathered, out of which 365 scene categories were chosen to be part of the dataset. Several annotators were asked to label every image and images with contradicting labels were discarded. Currently, the dataset is available in the Places365-standard format (i.e. 365 categories, roughly 1 million images training set, validation set with 50 images per class and test with 900 images per class) and the Places365-challenge format which extends the training set to 8 million image samples in total. With a dataset of this magnitude, the training of CNNs exclusively on data describing scenes becomes feasible.

\begin{table}[]
\caption{An overview of publicly available datasets for the task of scene recognition.}
\label{tab1}

\centering
\resizebox{\columnwidth}{!} {
\begin{tabular}{c|c|c|c|c|c|c}
\multirow{2}{*}{Dataset}  &  \multirow{2}{*}{Data}  & \multirow{2}{*}{ \#Classes } &  \multicolumn{2}{c|}{ Classification of } & \multicolumn{2}{c}{ Labelled as }  \\

& & & \ Images \ & \ Streams \ & \ Object \ & \ Scenes   \\ \hline
Places365 \cite{places365} & 1M images & 365 &\checkmark & & &\checkmark \\
MIT Indoor67 \cite{mitindoor67} & 15620 images & 67 &\checkmark & & &\checkmark  \\
SUN397 \cite{sun397} & 108754 images & 397 &\checkmark & & &\checkmark \\
15 scene \cite{sun397} & 4000 images & 15 &\checkmark & & &\checkmark  \\ \hline
Maryland `in-the-wild' \cite{maryland_dataset} & 10 videos & 13 & &\checkmark & &\checkmark  \\
YUPENN \cite{yupenn} & 410 videos & 14 & &\checkmark & &\checkmark  \\
YUP++ \cite{dynamic_scenes} & 1200 videos & 20 & & \checkmark& &\checkmark  \\ \hline
Imagenet \cite{imagenet} & 3.2M images & 1000 &\checkmark & &\checkmark & \\
COCO \cite{coco} & 1.5M images & 80 & \checkmark& &\checkmark &  \\
Open Images \cite{open_images} & 1.7M images & 600 &\checkmark & &\checkmark &  
\end{tabular}}
\end{table}

Scene recognition also encloses dynamic scene data; due to the limited amount of available datasets which include such data, most of the research efforts in this sub-field also include gathering suitable experimental data. Here we highlight the Maryland `in-the-wild' \cite{maryland_dataset}, YUPENN \cite{yupenn} , YUP++ \cite{dynamic_scenes} datasets. The dataset in \cite{dynamic_scenes} poses new challenges by introducing more complex data, i.e. videos with camera motion. The scope of the categories that are being recorded amongst the three datasets presented is not nearly as exhaustive as in the case of the objects and scenes datasets mentioned above. This is an indicator of the incipient status of research in this particular area of scene recognition.

The original models proposed by the authors of the \cite{maryland_dataset} and \cite{yupenn} datasets were not based on deep learning techniques. The authors of the the Maryland `in-the-wild' \cite{maryland_dataset}, introduced a chaotic system framework for describing the videos. The authors' proposed pipeline extracts a 960-dimensional Gist descriptor per videoframe. Each dimension is considered a time-series, from which the chaotic invariants are computed. Traditional classifiers, such as KNN and SVM, are used for the final classification. In \cite{yupenn}, the authors introduced the YUPENN dataset and for its analysis, they proposed a spatiotemporal oriented energy feature representation of the videos which they classify using KNN.

An overview of the described datasets is provided in Table \ref{tab1}. In Figure \ref{fig:places} we complete the quantitative overview of the datasets by presenting representative image samples for each of the datasets described.

\section{Frameworks for scene recognition}\label{sec:methods}
In this section, we describe relevant aspects of the state-of-the-art methods on scene recognition with deep learning. The choice for deep architectures is motivated by the complexity of the task: since the images are not described semantically the models used are aimed at learning generic contextual features of the scenes, which are captured by the high-level convolutional layers. 

Previous to deep learning, visual recognition techniques have made extensive use of object recognition when faced with such problems \cite{objects1,objects}. The scenes would be recognized based on exhaustive lists of objects identified in the scene. However, other challenges appear such as object detection and their high appearance variability. The combination of object detection and overall context recognition \cite{viswanathan2011place} showed promising results. 

Focusing on deep learning research papers, we group them based on the type of the analysed datasets, images or videos. We present their performances and limitations in the context of the evaluated datasets. 

\subsection{Static scene recognition}

Several works have addressed the recognition of scenes based on single image analysis. The best well-known work on scene recognition was introduced in \cite{places365}, which relied on the Places365 dataset. 

\begin{table}[h!]
\centering
\caption{Top-5 classification accuracy of the trained networks on the validation and test splits of the Places365 dataset. Apart from the ResNet architecture which has been fine-tuned over Places365, the other architectures are trained from scratch.}
\begin{tabular}{l|c|c}
Architectures  & \multicolumn{2}{c}{Top-5 accuracy}\\
trained on Places365 & \ Validation set \ & \ Test set \  \\
\hline \hline

Places365 AlexNet \cite{places365}&  82.89\% & 82.75\%  \\
Places365 GoogleNet\cite{places365} & 83.88\%  & 84.01\% \\
Places365 VGG \cite{places365}& 84.91\%  & 85.01\%    \\
Places365 ResNet \cite{places365} & 85.08\%& 85.07\%\\
\end{tabular}
    \label{tab:original_performance_places365}
\end{table}
Deep learning architectures have been trained over the Places365 dataset. The approach proposed by the authors of literature \cite{places365} is to exploit the vast dataset at hand by training three popular CNNs architectures (i.e. AlexNet \cite{alexnet}, GoogLeNet \cite{googlenet}, VGG16 \cite{vgg16}) on the Places dataset. The performance of these architectures over the validation and test splits of the Places365 dataset are presented in Table \ref{tab:original_performance_places365}. When introducing a new dataset, it became a ritual to test the generalization capabilities of weights trained over Places365. Thus, authors fine-tune these specialised networks trained on Places365 over newly available datasets. For instance, the VGG16\cite{vgg16}, pre-trained on the Places365 dataset, achieved a 92.99\% accuracy on the SUN Attribute dataset \cite{sun_database}. To compare the performance of the above approaches for static scene recognition, the following datasets are considered: 15 scenes dataset \cite{15_scenes}, MIT Indoor 67 \cite{mitindoor67} and SUN 397 \cite{sun397}. An overview of the comparison of the quantitative results is presented in Table \ref{tab:res_overview}.

\begin{table}
\caption{An overview of the quantitative comparison in terms of accuracy between methods for single image classification for the 15 scenes, MIT Indoor, SUN 397 datasets.} 
\centering
    \begin{tabular}{l| c| c| c}
    & 15 scenes & MIT Indoor & SUN 397  \\
    \hline 
    Places365 AlexNet \cite{places365}& 89.25\% & 70.72\% & 56.12\%   \\
    Places365 GoogleNet\cite{places365} & 91.25\%  & 73.20\%  & 58.37\%   \\
    Places365 VGG \cite{places365}& 91.97\%  & 76.53\%  & \textbf{63.24\%}   \\
    Hybrid1365 VGG \cite{places365} & \textbf{92.15\%}  & \textbf{79.49\%}  & 61.77\%   \\
    7-scale Hybrid VGG \cite{objects_scenes_scale} & \textbf{94.08\%}  & 80.22\% & 63.19\%*   \\
    7-scale Hybrid AlexNet \cite{objects_scenes_scale} & 93.90\%  & \textbf{80.97\%} & \textbf{65.38\%}   \\
    \end{tabular}
    \label{tab:res_overview}
\end{table}

Furthermore, in \cite{resnet152} the authors experimented with the ResNet152 residual network architecture, fine-tuned over the Places365. This work achieved a top-5 accuracy of 85.08\% and 85.07\% on the validation and, respectively, the test set of the Places365 dataset, as shown in Table \ref{tab:original_performance_places365}.



The use of the semantic and contextual composition of the image has been proposed by various approaches. For instance, in \cite{objscene}, the authors proposed the Hybrid1365 VGG architecture, a combination of deep learning techniques trained for object and scene recognition. The method uses different scales at which objects appear in a scene can facilitate the classification process by targeting distinct regions of interest within the image. Objects usually appear at lower scales. Therefore, the object classifier should target local scopes of the image. In contrast, the scene classifier should be aimed at the global scale, in order to capture contextual information. They concluded that it is possible to extend the performance obtained individually by each method. The Hybrid1365 VGG architecture \cite{objscene} scores the highest average accuracy of 81.48\% over all the experiments conducted for the place-centric CNN approach (has the highest performance for 2 out of 3 comparison datasets as shown in Table \ref{tab:res_overview}).
 
The dataset biases which arise under different scaling conditions of the images is addressed in \cite{objects_scenes_scale}, by involving a multi-scale model which combines various CNNs specialized either on object or place knowledge. The authors combined the training data available in the Places and ImageNet datasets. The knowledge learned from the two datasets is coupled in a scale-adaptive way. In order to aggregate the extracted features over the architectures used, simple max pooling\footnote{Max pooling is a pooling operation which computes the maximum value in each patch of a feature map; it is employed for down-sampling input representations.} is adopted in order to down-sample the feature space. If the scaling operation is significant, the features of the data can drastically change from describing scene data to object data. The architectures are employed to extract features in parallel from patches, which represent the input image at increasingly larger scale versions. The multi-scale model combines several AlexNet architectures \cite{alexnet}. The hybrid multi-scale architecture uses distinctive models for different scale ranges; depending on the scale range, the most suitable model is chosen from object-centric CNN (pre-trained on ImageNet), scene-centric CNN (pre-trained on Places365) or a fine-tuned CNN (adapted to the corresponding scale based on the dataset at hand). In total, seven scales were considered; the scales were obtained by scaling the original images between $227\times227$ and $1827\times1827$ pixels. For the final classification given by the multi-scale hybrid approach, the concatenation of the fc7 features (i.e. features extracted by the 7th fully connected layer of the CNN) from the seven networks are considered. Principal Component Analysis (PCA) is used to reduce the feature space. This model obtained the highest accuracy of 95.18\% on the 15 scenes dataset \cite{15_scenes}.

The hybrid approaches presented in \cite{objscene} and \cite{objects_scenes_scale} achieve higher accuracy than a human expert, which was quantified as 70.60\%. This indicates that the combination of object-centric and scene-centric knowledge can potentially establish a new performance standard for scene recognition.

\subsection{Dynamic scene recognition}
While early research in the field of scene recognition has been directed at single images, lately attention has been naturally drawn towards scene recognition from videos. CNNs have shown promising results for the general task of scene recognition in single images and have the potential to be also generalized to video data\cite{action2,conv_video}. To achieve this generalization, the spatio-temporal nature of dynamic scenes must be considered. While static scenes (depicted as single images) only present spatial features, videos also capture temporal transformations which affect the spatial aspect of the scene. Therefore, one challenge related to the task of scene classification from videos is creating a model which is powerful enough to capture both the spatial and temporal information of the scene. However, there are few works on video analysis for scene recognition. 

In the works introduced in \cite{bin2018describing,peris2016video}, the authors relied on Long Short Term Memory networks (LSTMs) for video description. However, they did not focus on recognizing the scenes.

\begin{table}[h!]
    \caption{Overview of the results achieved by the spatio-temporal residual network (T-ResNet) proposed in \cite{dynamic_scenes} over the YUP++ dataset.}
    \centering
    \begin{tabular}{l | c| c |c}
         & YUP++ static & YUP++ moving & YUP++ complete \\
         \hline  
        ResNet & 86.50\% & 73.50\% & 85.90\% \\
        T-ResNet & \textbf{92.41\%} & \textbf{81.50\%} & \textbf{89.00\%}
    \end{tabular}
    \label{tab:results_dynamic}
\end{table}

In \cite{dynamic_scenes}, the authors introduced the T-ResNet architecture, alongside the YUP++  dataset, which established a new benchmark in the sub-field of dynamic scene recognition. The T-ResNet is based on a residual network \cite{residual_nn_1} that was pre-trained on the ImageNet dataset \cite{imagenet}. It employs transfer learning to adapt the spatial-centric residual architecture to a spatio-temporal-centric network. The results achieved by the architecture were only compared with the classical ResNet architecture as shown in Table \ref{tab:results_dynamic}. 
The superiority of the T-ResNet is evident: it achieves an accuracy of 92.41\% on the YUP++ static camera partition, 81.50\% on the YUP++ moving camera partition and finally 89.00\% on the entire YUP++ dataset. This demonstrates the superiority of the spatio-temporal approach. The T-ResNet model exhibits strong performance for classes with linear motion patterns, e.g. classes `elevator', `ocean', `windmill farm'. However, for scene categories presenting irregular or mixed defining motion patterns the performance is negatively impacted, e.g. classes `snowing' and `fireworks'. The authors of \cite{dynamic_scenes} observed that T-ResNet exhibits difficulties distinguishing intrinsic scene dynamics from the additional motion of the camera. Further research is required to account for this difference.

\section{Discussion}
\label{sec:discussion}


The novel availability of large, exhaustive datasets, such as the Places Database, is offering significant support for further research for the challenge of scene recognition. The combination of scene-centric and object-centric knowledge has proven superior to only considering the scene context. Dynamic scene recognition reached new state-of-the-art performance through the approach of adapting spatial networks to the task, transforming the network to also consider the temporal aspect of the scenes. These emerging spatio-temporal networks are suitable for video data captured with a static camera. However, it still faces difficulties in the case of added camera motion.

One observation arising from methods addressing single image analysis scene recognition is that deeper CNN architectures such as GoogLeNet \cite{googlenet} or VGG \cite{vgg16} are not superior in all cases. For the hybrid multi-scale model combining scene-centric and object-centric networks in \cite{objects_scenes_scale}, experiments using VGG architecture for more than two-scales (two VGG networks) obtained disappointing results, inferior to the baseline performance achieved with one single scale (one network). Since the multi-scale hybrid model entails seven different scales, it can be inferred that VGG becomes noisy when applied on small input image patches.

Addressing the task of scene recognition from the global features that describe an image, the CNNs are expected to learn deep features that are relevant for the contextual clues present in the image. Literature \cite{places365} observers that the low-level convolutional layers detect low-level visual concepts such as object edges and textures, while the high-level layers activate on entire objects and scene parts. Even though the model has been previously trained on an exclusively places-centric dataset, the network still identifies semantic clues in the image by detecting objects alongside contextual clues. Therefore, CNNs trained on the Places Database (which does not contain object labels) could still be employed for object detection. 

Another aspect arising from training the same architecture on datasets with a different number of scene categories (i.e. and Places365) proves that having more categories leads to better results as well as more predicted categories. We can observe that the architecture AlexNet trained on Places205 (version prior to Places365) obtains 57.2\% accuracy, while the same architecture trained on Places365 obtains 57.7\% accuracy. For the places CNN approach two main types of miss-classifications occur: on one hand, less-typical activities happening in a scene context (e.g. taking a photo at a construction site) and on the other hand, images depicting multiple scene parts. A possible solution, as proposed by \cite{places365}, would be assigned multiple ground-truth labels in order to capture the content of an image more precisely.

The results achieved by the T-ResNet model illustrate the potential of spatio-temporal networks for video analysis. The transformation from a purely spatial network to a spatio-temporal one can succeed on the basis of a very small training set (i.e. only 10\% of the YUP++ dataset introduced) as proven by \cite{dynamic_scenes}. Well-initialized spatial networks can be efficiently transformed to extract spatio-temporal features, therefore, in theory, most networks that perform well on single image analysis could be easily adapted to video analysis.

\section{Conclusions}\label{sec:conclusion}

In this work, we describe the state-of-the-art on deep learning for scene recognition. Furthermore, we presented some of the applications of scene recognition to emphasize the importance of this topic. We argue that the main factor to consider is the type of data on which recognition and classification are applied. Since the task of scene recognition is not entirely subjective due to the nature of the scene images and the scene categories overlap, no one particular method can be generalized to all scene recognition tasks. This paper will aid professionals in making an informed decision about which approach best fits their scene recognition challenge. 
We have found room for research in the field of video analysis and expect that numerous works will emerge in the coming years.

\bibliographystyle{splncs04}
\bibliography{bibliography}

\end{document}